\documentclass[10pt,twocolumn,letterpaper]{article}

\usepackage{cvpr}
\usepackage{times}
\usepackage{epsfig}
\usepackage{graphicx}
\usepackage{amsmath}
\usepackage{amssymb}
\usepackage{color}
\usepackage{multirow}
\usepackage{subfigure}
\usepackage[abs]{overpic}

\usepackage[breaklinks=true,letterpaper=true,colorlinks,bookmarks=false]{hyperref}

\cvprfinalcopy 

\def\cvprPaperID{2630} 

\ifcvprfinal\fi
\begin{document}

\title{Iterative Learning with Open-set Noisy Labels}

\author{Yisen Wang\textsuperscript{1, 2}\ \  Weiyang Liu\textsuperscript{2}\ \  Xingjun Ma\textsuperscript{3}\ \  James Bailey\textsuperscript{3}\ \  Hongyuan Zha\textsuperscript{2}\ \ Le Song\textsuperscript{2}\ \ Shu-Tao Xia\textsuperscript{1}\\
  \textsuperscript{1}Tsinghua University \ \ \ \ \ \textsuperscript{2}Georgia Institute of Technology \ \ \ \ \ \textsuperscript{3}The University of Melbourne\\
{\tt\small wangys14@mails.tsinghua.edu.cn, wyliu@gatech.edu, xiast@sz.tsinghua.edu.cn }
}

\maketitle

\begin{abstract} 
Large-scale datasets possessing clean label annotations are crucial for training Convolutional Neural Networks (CNNs). However, labeling large-scale data can be very costly and error-prone, and even high-quality datasets are likely to contain noisy (incorrect) labels. Existing works usually employ a closed-set assumption, whereby the samples associated with noisy labels possess a true class contained within the set of known classes in the training data. However, such an assumption is too restrictive for many applications, since samples associated with noisy labels might in fact possess a true class that is not present in the training data. We refer to this more complex scenario as the \textbf{open-set noisy label} problem and show that it is nontrivial in order to make accurate predictions. To address this problem, we propose a novel iterative learning framework for training CNNs on datasets with open-set noisy labels. Our approach detects noisy labels and learns deep discriminative features in an iterative fashion. To benefit from the noisy label detection, we design a Siamese network to encourage clean labels and noisy labels to be dissimilar. A reweighting module is also applied to simultaneously emphasize the learning from clean labels and reduce the effect caused by noisy labels. Experiments on CIFAR-10, ImageNet and real-world noisy (web-search) datasets demonstrate that our proposed model can robustly train CNNs in the presence of a high proportion of open-set as well as closed-set noisy labels.
\end{abstract}

\section{Introduction}\label{sec:introduction}
The success of Convolutional Neural Networks (CNNs) \cite{krizhevsky2012imagenet} is highly tied to the availability of large-scale annotated datasets, \eg, ImageNet \cite{deng2009imagenet}. However, large-scale datasets with high-quality label annotations are not always available for a new domain, due to the significant time and effort it takes for human experts. There exist several cheap but imperfect surrogates for collecting labeled data, such as crowd-sourcing from non-experts or annotations from the web, especially for images (\eg, extracting tags from the surrounding text or query keywords from search engines). These approaches provide the possibility to scale the acquisition of training labels, but invariably result in the introduction of some noisy (incorrect) labels. Moreover, even high-quality datasets are likely to have noisy labels, as data labeling can be subjective and error-prone. The presence of noisy labels for training samples may adversely affect representation learning and deteriorate prediction performance \cite{nettleton2010study}. Training accurate CNNs against noisy labels is therefore of great practical importance.

\begin{figure}[!t]
\centering
\includegraphics[width=0.85\linewidth]{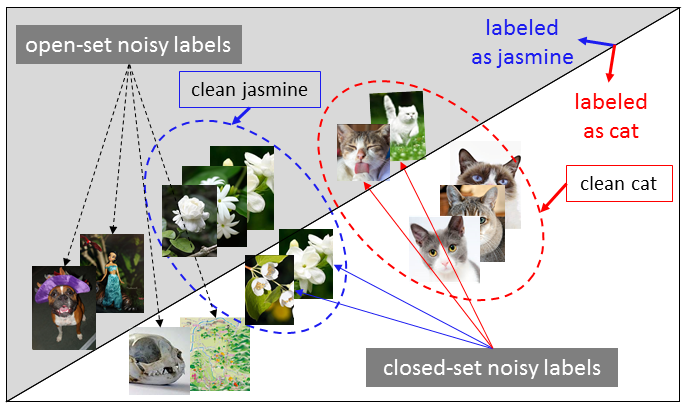}
\caption{An illustration of closed-set vs open-set noisy labels. }
\vspace{-0.1 in}
\label{fig:openset_label_noise}
\end{figure}

\begin{figure}[!t]
\centering
\includegraphics[width=0.9\linewidth]{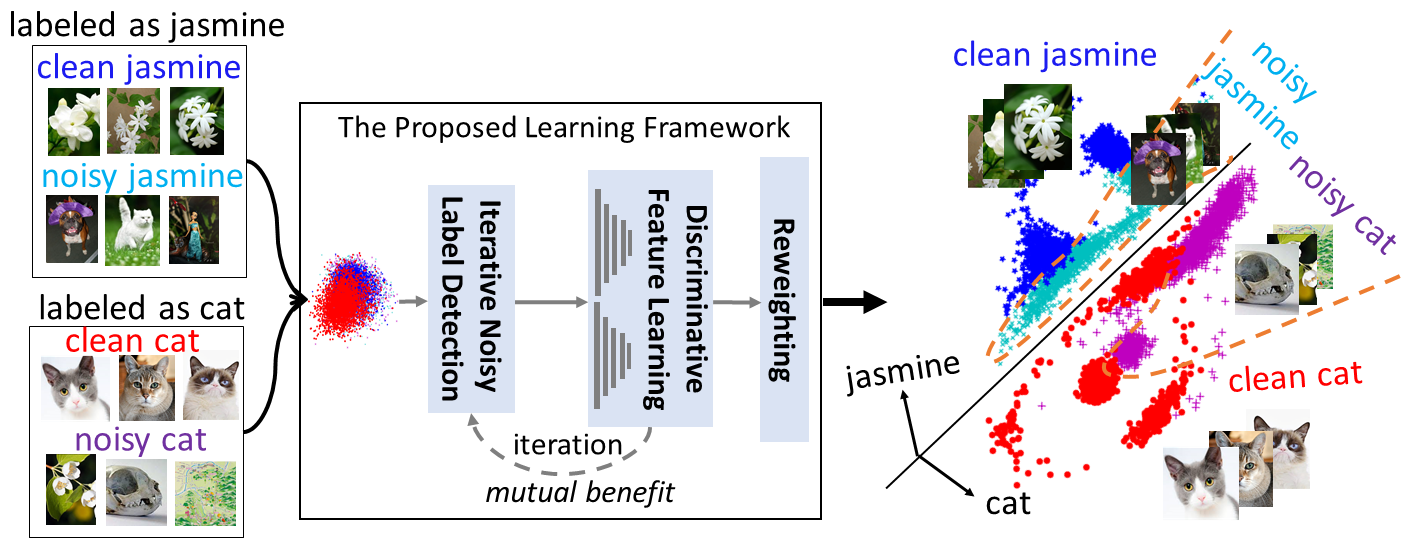}
\caption{An overview of our framework that iteratively learns discriminative representations on a ``jasmine-cat" dataset with open-set noisy labels. It not only learns a proper decision boundary (the black line separating jasmine and cat) but also pulls away noisy samples (green and purple) from clean samples (blue and red).}
\vspace{-0.1 in}
\label{fig:overview}
\end{figure}

We will refer to samples whose classes are mislabeled/incorrectly annotated as {\em noisy samples} and denote their labels as {\em noisy labels}.   Such noisy labels can fall into two types, {\em closed-set} and {\em open-set}. More specifically, a {\em closed-set noisy label} occurs when a noisy sample possesses a true class that is contained within the set of known classes in the training data. While, an {\em open-set noisy label} occurs  when a noisy sample possesses a true class that is not contained within the set of known classes in the training data. The former scenario has been studied in previous work, but the latter one is a new direction we explore in this paper. Figure~\ref{fig:openset_label_noise} provides a pictorial illustration of noisy labels, where we have an image dataset with two classes, jasmine (the plant) and cat (the animal). The closed-set noisy labels occur when cat and jasmine are mislabeled from one category to the other, but the true labels of these images are still cat or jasmine. The open-set noisy labels occur for those images labeled as cat or jasmine, but their true labels are neither cat nor jasmine, \eg, the zoo map and the cartoon character. Table \ref{tab:noise_type} demonstrates all the possible cases on how different samples are labeled in this problem. The leftmost column specifies the true class and the other columns specify the type of label in the dataset. 

\begin{table}[!t]
\centering
\small
\caption{Types of labels for a ``jasmine-cat" dataset.}
\label{tab:noise_type}
\vspace{0.05in}
\begin{tabular}{l|c|c}
& labeled as ``jasmine" & labeled as ``cat" \\ \hline
true ``jasmine" & clean & closed-set  \\ \hline
true ``cat" & closed-set  & clean  \\ \hline
other class images & open-set & open-set 
\vspace{-0.1 in}
\end{tabular}
\end{table}

Previous work has addressed the noisy label problem explicitly or implicitly in a closed-set setting, via either loss correction or noise model based clean label inferring \cite{Li_2017_ICCV,Patrini_2017_CVPR,vahdat2017toward,VeitACKGB17}. However, these methods are vulnerable in the more generic open-set scenario, as loss or label correction may be inaccurate since the true class may not exist in the dataset. Open-set noisy labels are likely to occur for scenarios where data are harvested rapidly, or use approximate labels (\eg, using a search engine query to retrieve images and then labeling the images according to the query keyword that was used). To the best of our knowledge, how to address the open-set noisy label problem is a new challenge.

In this paper, we propose an iterative learning framework that can robustly train CNNs on datasets with open-set noisy labels. Our model works iteratively with:
(1) a noisy label detector to iteratively identify noisy labels; 
(2) a Siamese network for discriminative feature learning, which imposes a representation constraint via contrastive loss to pull away noisy samples from clean samples in the deep representation space; and (3) a reweighting module on the softmax loss to 
express a relative confidence of clean and noisy labels on the representation learning. A simplified illustration of the proposed framework is presented in Figure~\ref{fig:overview}. Our main contributions can be summarized as follows:

(1) We identify the open-set noisy label problem as a new challenge for representation learning and prediction. 

(2) We propose an iterative learning framework to robustly train CNNs in the presence of open-set noisy labels. Our model is not dependent on any assumption of noise.

(3) We empirically demonstrate that our model significantly outperforms state-of-the-art noisy label learning models for the open-set setting, and has a comparable or even better performance under the closed-set setting.


\section{Related work}
A simple approach to handle noisy labels is to remove samples with suspicious labels from the training data \cite{BrodleyF99}. However, such methods are often challenged by the difficulty of distinguishing samples that are inherently hard to learn from those with noisy labels \cite{guyon1996discovering}.  In contrast to simply removing them, the following work focuses on addressing the noisy label problem via deep learning.

One alternative approach is to explicitly or implicitly formulate the noise model and use a corresponding noise-aware approach. Symmetric label noise that is independent of the true label was modeled in \cite{larsen1998design}, and asymmetric label noise that is conditionally independent of the individual sample was modeled in \cite{NatarajanDRT13, sukhbaatar2014training}. More complex noise models for samples, true labels and noisy labels can be characterized by directed graphical models \cite{XiaoXYHW15}, Conditional Random Fields (CRF) \cite{vahdat2017toward}, neural networks \cite{VeitACKGB17} or knowledge graphs \cite{Li_2017_ICCV}. These methods aim to correct noisy labels to their true labels via a clean label inferring. However, they require availability of an extra dataset with pre-identified noisy labels and their ground truth labels in order to model label noise. Moreover, these methods make their own specific assumptions about the noise model, which will limit their effectiveness under complicated label noise.

Other approaches utilize correction methods to adjust the loss function to eliminate the influence of noisy samples. Backward \cite{Patrini_2017_CVPR} and Forward \cite{Patrini_2017_CVPR} are two such correction methods that use an estimated or learned factor to modify the loss function. \cite{sukhbaatar2014learning, goldberger2016training} further augment the correction architecture by adding a linear layer on top of the network.  Bootstrap \cite{reed2014training} is another loss correction method that replaces the target labels used by a loss function with a combination of raw target labels and their predicted labels. 

The above methods implicitly assume a closed-set noisy label setting, where the true labels are always contained within the set of known classes in the training data. Such restricted assumption contradicts the more practical open-set scenario. Open-set noisy samples should be considered separately. In our proposed model, we iteratively detect noisy samples and gradually pull them away from clean samples, which is different from removing them or labeling them to a new ``unknown" class \cite{sukhbaatar2014training} (these two approaches are evaluated in Section \ref{sec: module analysis}). Moreover, the proposed framework does not depend on the noise model, and is able to address both the open-set and the closed-set noisy label problem.


\section{Iterative learning framework}

\begin{figure*}[!t]
\centering
\includegraphics[width=0.8\linewidth]{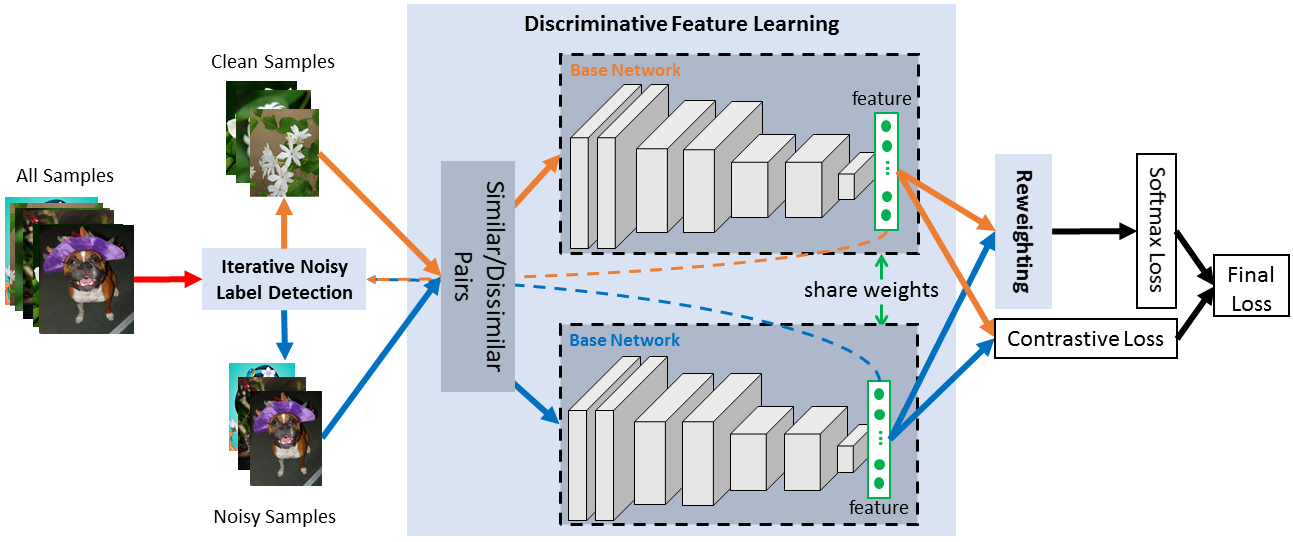}
\caption{The framework of the proposed iterative learning approach. Iterative noisy label detection module and discriminative feature learning module form a closed-loop, \textit{i.e.}, one module's inputs are the other module's output, which can benefit from each other and be jointly enhanced. The network is jointly optimized by two types of losses: reweighted softmax loss and contrastive loss. }
\vspace{-0.1in}
\label{fig:framework}
\end{figure*}

Our goal is to learn discriminative features from a dataset with noisy labels. We propose an iterative learning framework that gradually pulls away noisy samples from clean samples in the deep feature space. As illustrated in Figure~\ref{fig:framework}, our proposed model consists of three major modules: 1) iterative noisy label detection, 2) discriminative feature learning, and 3) reweighting. The noisy label detection uses the output features of the network (dashed lines) to separate training samples into two subsets: clean samples and noisy samples. To benefit from the noisy label detection, we employ a Siamese network to impose a representation constraint forcing the representation of clean samples and that of noisy samples to be as discriminative as possible. Besides, a reweighting module that assigns a weight for each sample based on the confidence supplied by the noisy label detection is used to emphasize clean samples and weaken noise samples on the discriminative representation learning. Such learned discriminative representations will in turn benefit the noisy label detection. Considering the representation learning as an iterative process, we further design the noisy label detection to be iterative so that the discriminative feature learning and the iterative noisy label detection can be jointly improved over iterations. A brief description of how each module works is listed as follows:
{\bf Iterative noisy label detection:} 
We iteratively detect noisy labels based on the features of the network, because samples from the same class should be intrinsically similar, while mislabeled samples are generally not \cite{bouveyron2009robust}. 

{\bf Discriminative feature learning:} 
We use a Siamese network with two sub-networks of sharing weights. It takes ``similar" or ``dissimilar" sample pairs as inputs and uses a contrastive loss to minimize distance between similar samples and maximize distance between dissimilar samples. It can also be seen as a representation constraint. 

{\bf Reweighting:} For detected clean samples, we set their weights to 1 (no reweighting) on softmax loss, while for detected noisy samples, we assign them smaller weights individually based on how likely one sample being noisy. To avoid misdetection, samples near the decision boundary will be weighted of close importance to clean samples.

The framework is jointly optimized by two loss terms:
\begin{equation}
L = {\rm RSL} + \eta {\rm CL},
\end{equation}
where RSL is the reweighted softmax loss, CL is the contrastive loss and $\eta$ is a trade-off parameter. The above objective incorporates the iterative noisy label detection, discriminative feature learning and reweighting into an effective learning framework that is robust to noisy labels. 

\subsection{Iterative noisy label detection}
\label{sec:iterative_detection}
Considering that samples from the same class should have similar high-level representations but samples mislabeled into the class do not \cite{bouveyron2009robust}, we detect noisy labels based on the representations of the pre-softmax layer. To benefit from the iterative learning process of representation, we iteratively perform noisy label detection every few epochs. We also use a cumulative criterion based on all previous iterations of detection to reduce the influence of randomness in one particular iteration and further produce more stable detection results. 

Our detection method is a probabilistic and cumulative version of Local Outlier Factor algorithm (pcLOF), which inherits the advantages of LOF, \textit{i.e.}, it is an unsupervised algorithm which performs well on high dimensional data and requires no assumptions of the underlying data distribution. Formally, pcLOF is defined as:
\begin{equation}
{\rm pcLOF}(x_i) = G\Big(\sum_{m = 1}^M {\rm LOF}^{(m)}(x_i) \Big), 
\end{equation}
where $M$ is the current number of iteration and $G$ is a local Gaussian statistics transformation, which scales the cumulative LOF score to a probabilistic value in [0, 1] as in \cite{kriegel2009loop,kriegel2011interpreting}. The pcLOF score can be directly interpreted as the probability of a sample being an outlier. In the noisy label detection setting, a pcLOF score close to 0 indicates a clean sample, while a score close to 1 indicates a noisy sample.

LOF is a density-based outlier detection algorithm \cite{BreunigKNS00} and the LOF score of a sample $x_i$ is defined as follows:
\begin{equation}
{\rm LOF}(x_i) = \frac{\sum_{x_j \in N_k(x_i)} \frac{lrd(x_j)}{lrd(x_i)}}{|N_k(x_i)|},
\end{equation}
where $N_{k}(x_i)$ is the set of $k$ nearest neighbors of $x_i$ and $lrd(x_i)$ is the local reachability density ($lrd$) of $x_i$:
\begin{equation}
lrd(x_i) = 1/ \bigg(\frac{\sum_{x_j \in N_k(x_i)}  \textit{reach-dist}_{k}(x_i, x_j)}{|N_k(x_i)|}\bigg), 
\end{equation}
where $\textit{reach-dist}_{k}(x_i,x_j)=\max \{\textit{k-dist}(x_j),d(x_i,x_j)\}$ is the reachability distance of $x_i$ to $ x_j$. Intuitively, if $x_i$ is far away from $x_j$, then the reachability distance is simply $d(x_i,x_j)$ (their actual distance). However, if they are ``sufficiently" close, the actual distance is replaced by $\textit{k-dist}(x_j)$ (the distance of $x_j$ to its $k$-th nearest neighbor), which means that samples inside of the $k$ nearest of $x_j$ are considered to be equally distant. 

Note that the noisy label detection works iteratively, thus we do not need complicated detection algorithms. With the representation become more discriminative, 
they can converge to almost the same result as long as the iteration is long enough. To balance the efficiency and effectiveness in training, we perform pcLOF based iterative noisy label detection every 10 epochs after 2-epoch network initialization in our experiments.

\subsection{Discriminative feature learning}
\label{sec:contrastive_loss}
We implement a Siamese network \cite{chopra2005learning, HadsellCL06} with two channels of the same base network and sharing weights. It generates ``similar" and ``dissimilar" sample pairs based on the clean and noisy samples detected by the noise label detection module, and works with a contrastive loss to minimize distance between samples of the same class and maximize distance between samples of different classes as well as distance between clean samples and noisy samples.

Denote the Euclidean distance between $x_i$ and $x_j$ in the deep representation space as follows:
\begin{equation}
\mathcal{D}_l(x_i, x_j, \theta) = ||f^l(x_i | \theta) - f^l(x_j | \theta) ||_2,
\end{equation}
where $f^l(\cdot | \theta)$ denotes the $l$-th layer output of the network $f(\cdot| \theta)$ under parameters $\theta$. The similarity indicator $Y_{ij}$ is defined based on the output of noisy label detection: 
\begin{equation}
Y_{ij} = 
\begin{cases}
1, &  \text{if $x_i$ and $x_j$ are similar}; \\
0, &  \text{if $x_i$ and $x_j$ are dissimilar}.
\end{cases}
\end{equation}
Two samples are considered to be ``similar" ($Y_{ij} = 1$), if and only if the two samples are from the same class and both are correctly labeled. Two samples are considered to be ``dissimilar" ($Y_{ij} = 0$), if two samples are from different classes, or one is a clean sample and the other is a noisy sample. Note that we do not define the relationship between two noisy samples as their true classes are not contained within the training data, thus cannot be simply defined as similar or dissimilar. When noisy samples are not available before the first iteration of noisy label detection, dissimilar pairs only contain the samples from different classes. 

The contrastive loss for discriminative feature learning can be formulated as:
\begin{equation}
{\rm CL}(x_i, x_j, Y_{ij}) = Y_{ij} \frac{1}{2} \mathcal{D}_l^2 +  (1-Y_{ij}) \frac{1}{2} \max \{0, \alpha - \mathcal{D}_l\},
\end{equation}
where $\alpha > 0$ is a margin formulating how far away two dissimilar samples should be pulled from each other. This contrastive loss will force the distance between dissimilar pairs, \textit{e.g.}, clean samples and noisy samples, to be larger than the margin $\alpha$, and similar pairs, \textit{i.e.}, clean samples from the same class, to be clustered.

Although the amount of possible sample pairs can be huge, some of the pairs are easy to discriminate and ineffectual for training (\textit{i.e.}, the distance between two dissimilar samples is already larger than the margin $\alpha$). Therefore, we implement the widely used hard example mining strategy \cite{simo2015discriminative} to obtain the closest dissimilar pairs and the most distant similar pairs to feed into the network. The base network can be any kind of architectures, such as VGG \cite{simonyan2014very}, ResNet \cite{he2016deep} and Inception \cite{szegedy2016rethinking}. 

\subsection{Reweighting}
\label{sec:reweighting}
To ensure an efficient and accurate representation learning, we also design a reweighting module before the softmax loss to adaptively use the label information with different confidence. Applying softmax loss on clean samples is intuitively to make use of their trustworthy label information. The reason for also applying softmax loss on noisy samples is that the detected noisy samples may contain some clean samples, especially at the start of the training. Those samples are close to the decision boundary and are often very informative for representation learning as pointed out in \cite{guyon1996discovering}, which cannot be simply ignored, \textit{i.e.}, setting their weights to 0. This is also verified by the experiments in Section \ref{sec: module analysis}. Therefore, a reweighting module paired with the softmax loss is used to adaptively address the detected clean and noisy samples simultaneously.

Since the noisy label detection method pcLOF provides the confidence/probability of a sample being noisy, we use $\gamma = 1 - {\rm pcLOF}$ as a reweighting factor on the detected noisy samples to express their relative confidence. 
We set the initial $\gamma$ to 1 before the first iteration of noisy label detection, as the learning proceeds, the $\gamma$ of a noisy sample will be gradually decreased as the noisy label detection tends to be more and more accurate. The reweighted softmax loss (RSL) of our model is defined as:
\begin{equation}
\small
\begin{split}
{\rm RSL} & = - \frac{\Big( \sum_{i = 1}^{N_c} \log P(y_i|x_i, \theta) + \sum_{j = 1}^{N_n} \gamma_j \log P(y_j|x_j, \theta) \Big)}{N_c + N_n},
\end{split}
\end{equation}
where $N_c$, $N_n$ denote the number of detected clean and noisy samples respectively, $\gamma$ is the proposed reweighting factor on the detected noisy samples, and $P(y_i|x_i, \theta)$ is the softmax probability of $x_i$ being in class $y_i$. The softmax loss used in our framework can also be replaced to more advanced ones \cite{liu2016large,liu2017sphereface} to further boost the performance.

\section{Experiments}\label{sec:experiments}
In this section, we evaluate the robustness of our proposed model to noisy labels with comprehensive experiments on  CIFAR-10 (small dataset),   ImageNet (large-scale dataset), and web-search dataset (real-world noisy dataset). 

\subsection{Exploratory experiments on CIFAR-10}\label{sec:small_datasets}
We first conduct a series of experiments on CIFAR-10 dataset towards a comprehensive understanding of our model through comparisons to the state-of-the-arts. 

\noindent\textbf{Baselines:} 
Several recently proposed noisy label learning models are chosen as our baselines: (1) Backward \cite{Patrini_2017_CVPR}: The networks are trained via loss correction by multiplying the cross entropy loss by an estimated correction matrix; (2) Forward \cite{Patrini_2017_CVPR}: The networks are trained with label correction by multiplying the network prediction by an estimated correction matrix; (3) Bootstrap \cite{reed2014training}: The networks are trained with new labels generated by a convex combination (the ``hard" version) of the noisy labels and their predicted labels; and (4) CNN-CRF \cite{vahdat2017toward}: The networks are trained with latent clean labels inferred by CRF from only noisy training datasets. We also include the method of Cross-entropy: learning directly from noisy datasets with a vanilla cross entropy loss.

\noindent\textbf{Experimental setup:}
The baseline models use a network architecture with 6 convolutional layers and 1 fully connected layer ({\em fc7}). Batch normalization (BN) \cite{ioffe2015batch} is applied in each convolutional layer before the ReLU activation, a max-pooling layer is implemented every two convolutional layers, and a softmax layer is added on top of the network for classification. The parameters of the baselines are configured according to their original papers. For our model, two copies of the above network are implemented and the contrastive loss is built upon the {\em fc7} layer. We set $\eta = 1$, $k = \textit{half the class sample size}$, and samples with ${\rm pcLOF} > 0.5$ are considered as noisy samples\footnote{As pointed by \cite{kriegel2009loop,kriegel2011interpreting}, pcLOF is not sensitive to parameter $k$}. 
The classification accuracy (ACC) on clean CIFAR-10 test set is used as the evaluation metric. 

All networks are trained by Stochastic Gradient Descent (SGD) with learning rate 0.01, weight decay $10^{-4}$ and momentum 0.9, and the learning rate is divided by 10 after 40 and 80 epochs (100 in total). All images are mean-subtracted and normalized to $[0,1]$, and no data augmentation is implemented in this part.

\begin{figure}[!t]
\centering
\includegraphics[width=0.95\linewidth]{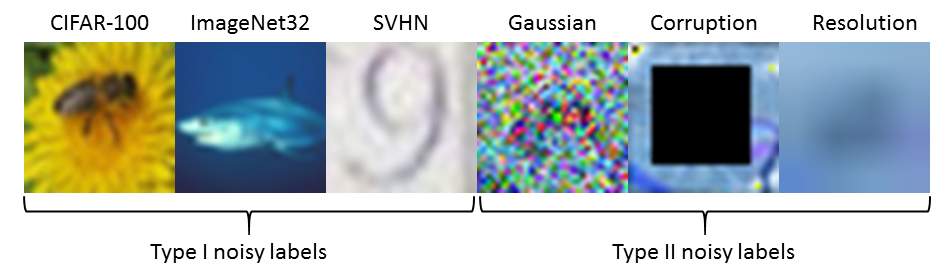}
\caption{Examples of open-set noise for ``airplane" in CIFAR-10.}
\vspace{-0.1 in}
\label{fig:noise_type}
\end{figure}

Open-set noisy datasets are built by replacing some training images in CIFAR-10 by outside images, while keeping the labels and the number of images per class unchanged. The ``mislabeled" outside images are from either different public datasets (type I noise) or severely damaged CIFAR-10 images (type II noise). Type I noise includes images from CIFAR-100 \cite{krizhevsky2009learning}, ImageNet32 (32$\times$32 ImageNet images) \cite{chrabaszcz2017downsampled} and SVHN \cite{netzer2011reading}, and only those images whose labels exclude the 10 classes in CIFAR-10 are considered. Type II noise includes images damaged by Gaussian random noise (mean 0.2 and variance 1.0), corruption ($75\%$ of an image is set to black or white) and resolution distortion 
(an image is resized to 4$\times$4 and then dilated back to 32$\times$32). Some examples of the type I and type II open-set noise are given in Figure~\ref{fig:noise_type}. For closed-set noisy datasets, we choose a random proportion of CIFAR-10 training images per class and change its label to an incorrect random one. This closed-set label noise belongs to symmetric noise, which is more challenging than asymmetric noise \cite{Patrini_2017_CVPR}. 

\begin{table*}[!htb]
\renewcommand{\arraystretch}{1.1}
\centering
\small
\caption{Accuracies (\%) of different models on CIFAR-10 noisy dataset with $40\%$ open-set noise. The best results are in \textbf{bold}.}
\label{cifar10_openset}
\vspace{0.05in}
\begin{tabular}{ll|cccccc}
\hline
\multicolumn{2}{c|}{Open-set label noise type} & Cross-entropy & Backward & Forward & Bootstrap & CNN-CRF & Ours  \\ \hline
\multirow{3}{*}{Type I} & CIFAR-10 + CIFAR-100 & 62.92 & 55.97 & 64.18 & 62.11 & 64.58 & \textbf{79.28}  \\
& CIFAR-10 + ImageNet32 & 58.63 & 52.35 & 66.77 & 57.83 & 67.53 & \textbf{79.38}  \\
& CIFAR-10 + SVHN & 56.44  & 52.03 & 56.70  & 56.89 & 56.93 & \textbf{77.73}  \\ \hline
\multirow{3}{*}{Type II} & CIFAR-10 + Gaussian & 61.96 & 54.98 & 72.70 & 59.05 & 72.51 & \textbf{80.37}  \\
& CIFAR-10 + Corruption & 57.40  & 50.24 & 63.80 & 56.00 & 64.25 & \textbf{74.48}  \\
& CIFAR-10 + Resolution & 56.93  & 49.58 & 62.65 & 58.95 & 63.60 & \textbf{77.30}  \\
\hline
\end{tabular}
\end{table*}

\begin{table}[!htb]
\renewcommand{\arraystretch}{1.1}
\centering
\small
\caption{Accuracies (\%) on CIFAR-10 noisy dataset with $20\%$ and $40\%$ closed-set noise. Top 2 results are in \textbf{bold} except for ``Clean''. }
\vspace{0.05in}
\label{cifar10_closeset}
\begin{tabular}{l|cc}
\hline
Method & $20\%$ noise  & $40\%$ noise \\ \hline
Clean & 84.85 & 84.85 \\ 
Cross-entropy & 74.17 & 62.38 \\ \hline
Backward & 76.27 & 75.18 \\
Forward & 79.25 & 77.81  \\
Bootstrapping & 74.39 & 69.50 \\
CNN-CRF & \textbf{80.15} & \textbf{78.69} \\\hline
Ours & \textbf{81.36} & \textbf{78.15} \\ \hline
\end{tabular}
\end{table}

\subsubsection{Classification performance}
\label{classification_small}

\noindent\textbf{Open-set label noise:} 
The classification accuracy on CIFAR-10 noisy datasets with $40\%$ open-set noise is reported in Table \ref{cifar10_openset}. As can be seen, our model outperforms the baselines with large margins on both type I (top three rows) and type II (bottom three rows) open-set noise. The poor performance of baselines is because they either ignore the existence of noisy labels such as Cross-entropy, or attempt to correct noisy labels to so-called ``clean labels". For example, in CIFAR-10+SVHN, noisy images from SVHN are still noisy even if their labels are corrected to one of the CIFAR-10 classes, thus still harm representation learning. 
Our superior performance indicates that our model is capable of learning accurate representation directly from datasets with open-set noisy labels. Such capability opens up more opportunities for many applications that, for example, require learning directly from web-search data.

\noindent\textbf{Closed-set label noise:} 
We also assess our model under the closed-set noise settings of $20\%$ and $40\%$ noise rates. The results are reported in Table \ref{cifar10_closeset}. The first row ``clean" provides a performance ``upper bound" trained on the totally clean CIFAR-10 dataset. Compared to baselines, our model achieves comparable or better performance. In particular, our model surpasses all baselines at the 20\% noise level and its performance is close to the clean training. At the 40\% noise level, our model achieves an accuracy that is comparable to CNN-CRF and higher than other baselines. This experiment demonstrates that our model can also effectively learn from datasets with closed-set noisy labels. 

\noindent\textbf{Discussion:} Revisiting Table \ref{cifar10_openset}, we find that some baselines (Forward and CNN-CRF) achieve considerable improvements on CIFAR-10+CIFAR-100 and CIFAR-10+ImageNet32, compared to Cross-entropy. This may be caused by the similarity between images from CIFAR-100/ImageNet32 and those in CIFAR-10. Since CIFAR-100 and ImageNet32 have many fine-grained classes, some noisy images from these two datasets can be regarded as the closed-set noise rather than the open-set noise. However, for dissimilar datasets, \eg, CIFAR-10+SVHN, we can see that those baselines perform almost the same poorly. This interesting finding implies another evidence that our model still works well when open-set and closed-set noise coexist.

\subsubsection{Model interpretation}
We further demonstrate some visual results to help understand how our model works. 

\begin{figure}[!htbp]
\centering
\vspace{-0.25in}
\includegraphics[width=0.8\linewidth]{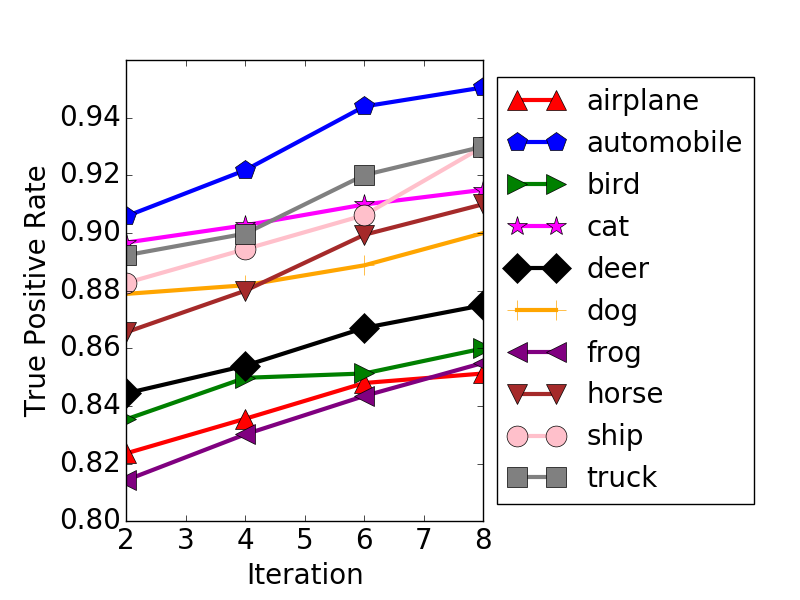}
\vspace{-0.06in}
\caption{The true positive rate of the detected noisy labels over iteration on CIFAR-10+CIFAR-100 (40\% open-set noise).}
\label{detection_rate}
\end{figure}

\begin{figure}[!t]
\centering
\small
\includegraphics[width=0.8\linewidth]{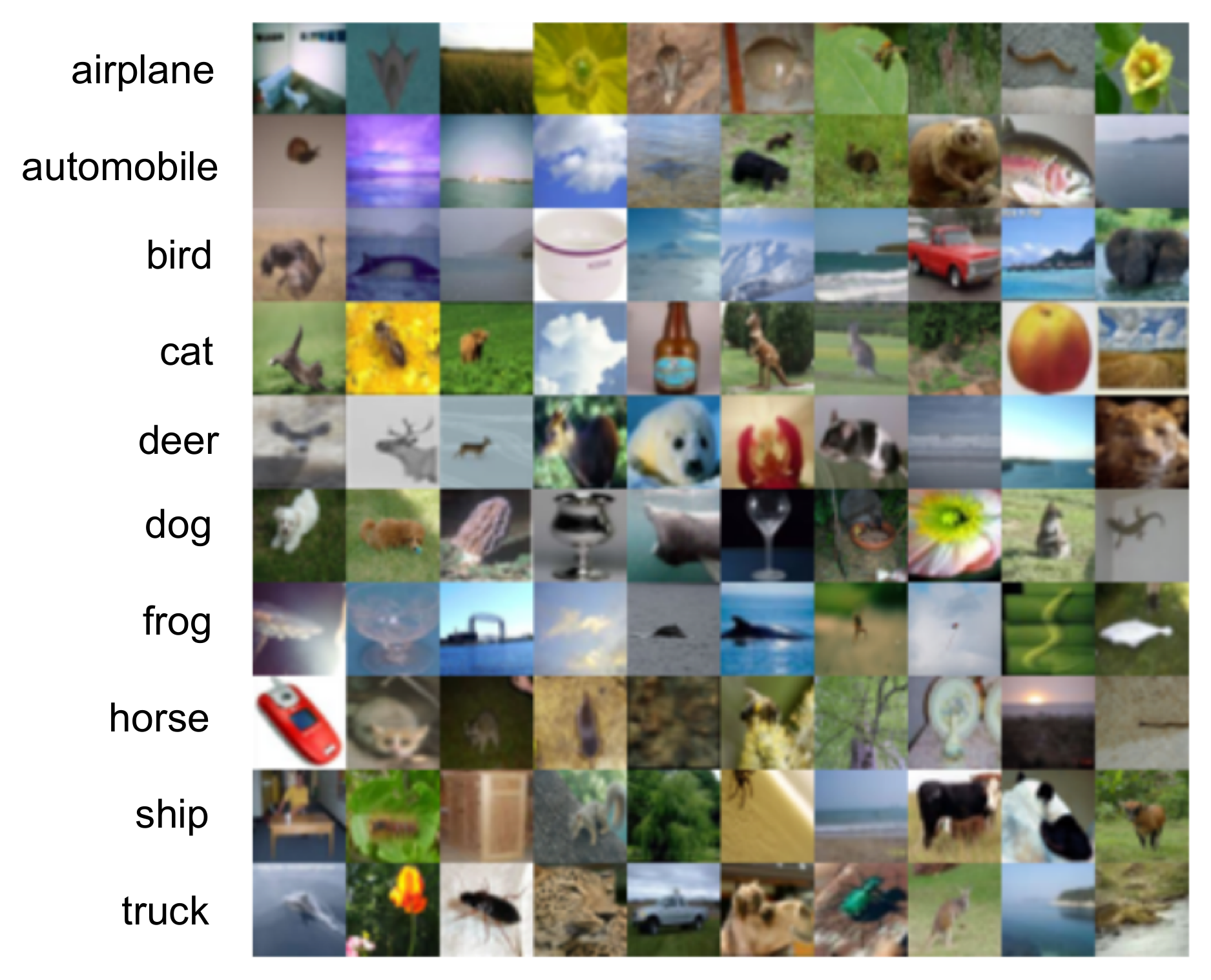}
\vspace{-0.05in}
\caption{Randomly selected images (10 per class) from the detected noisy images of CIFAR-10+CIFAR-100 with $40\%$ open-set noise at $100$-th epoch.}
\vspace{-0.1in}
\label{fig:outlier_visual}
\end{figure}

\noindent\textbf{Iterative noisy label detection:}
We first show the effectiveness of the iterative noisy label detection module. We use the measure of true noisy label rate (true positive rate) of the detected noisy labels. In Figure \ref{detection_rate}, the detection becomes more and more precise as the training proceeds, and the trend is consistent across tall classes, which meets our expectation that the iterative noisy label detection will improve accordingly as the learned features become increasingly discriminative. Figure \ref{fig:outlier_visual} gives some of those detected noisy images. We can see many noticeable open-set noisy images, \eg, flower, fish to name a few, which confirms that our iterative noisy label detection method can indeed accurately identify noisy samples contained within the dataset.

\begin{figure*}[!t]
\centering
\small
\includegraphics[width=0.24\linewidth]{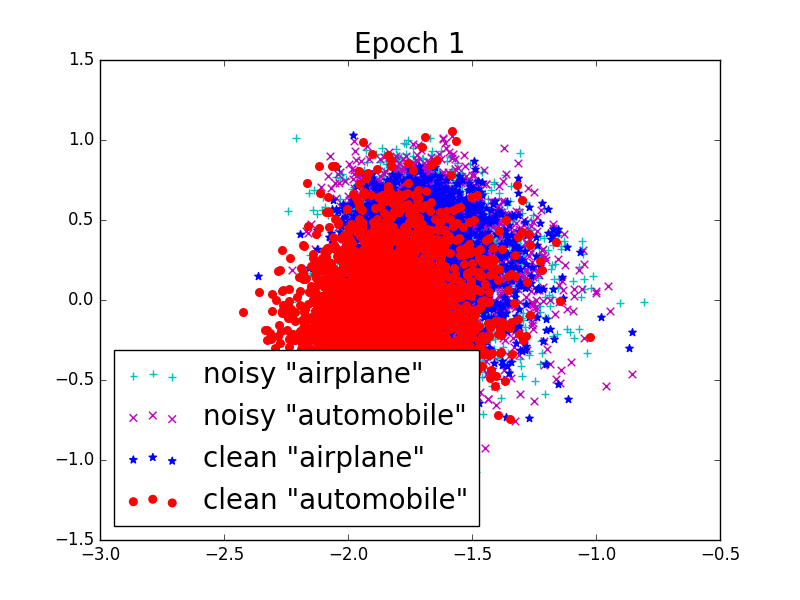}
\includegraphics[width=0.24\linewidth]{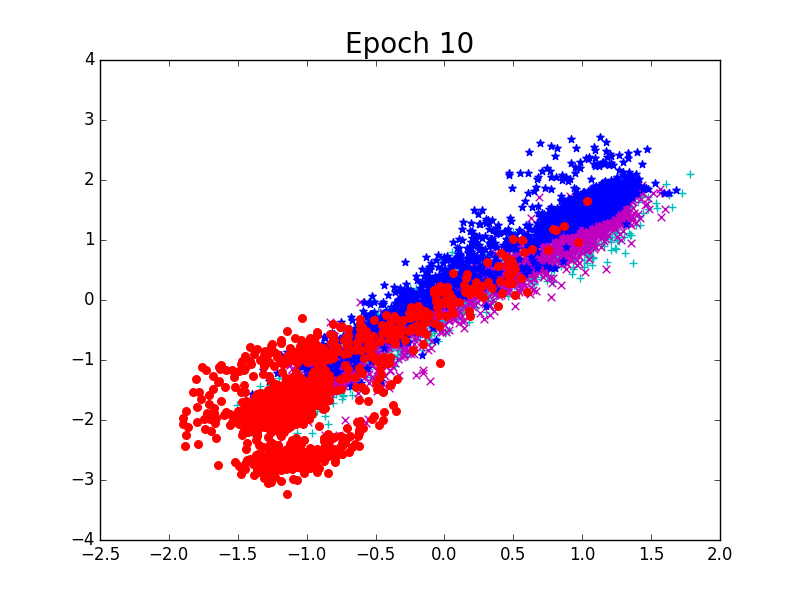}
\includegraphics[width=0.24\linewidth]{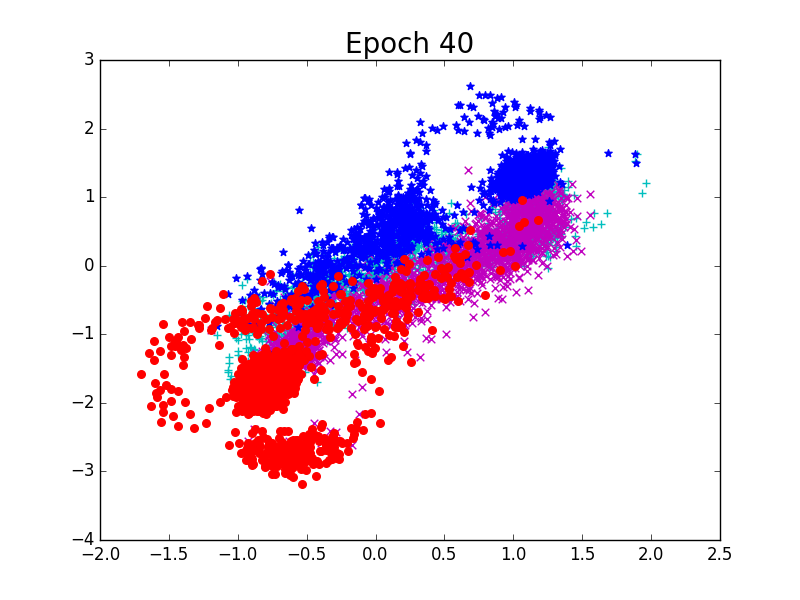}
\includegraphics[width=0.24\linewidth]{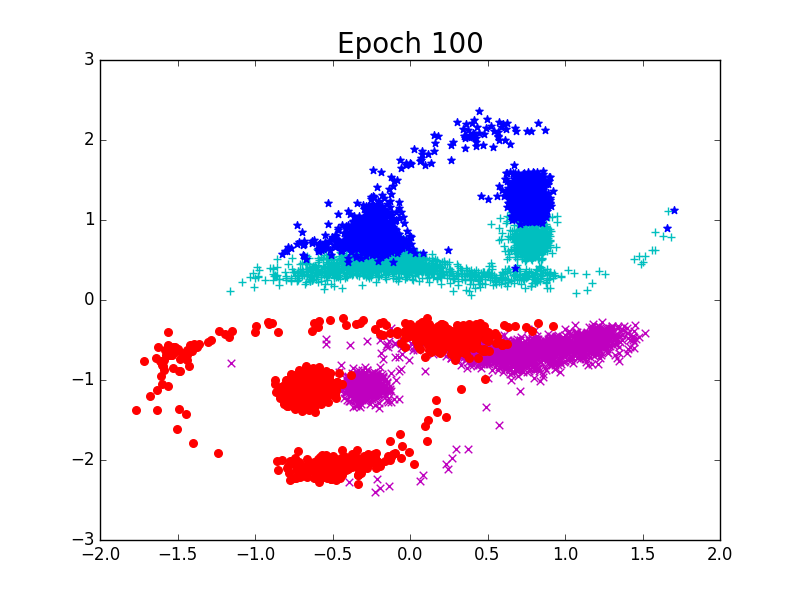}
\caption{Visualization of the learned features. This visualization experiment uses a 2-class subset of CIFAR-10+CIFAR-100 (40\% open-set noise) by setting the output feature dimension as 2.}
\vspace{-0.1in}
\label{representation_learning}
\end{figure*}

\noindent\textbf{Discriminative features:}
Next, we show the discriminative feature learning ability of our model by visualizing the learned features at different stages of the training process in 2-D space. Figure \ref{representation_learning} evidently shows that the learned features become more and more discriminative as noisy samples are gradually pulled away from clean samples. At the first epoch, all samples are densely overlapped together, as the training proceeds to the $100$-th epoch, not only the two classes have been separated, the noisy samples are also pulled away from the clean ones within each class. This confirms that the discriminative feature learning module can work effectively with the iterative noisy label detection module to isolate noisy samples. 

\subsubsection{Module analysis via ablation experiments} \label{sec: module analysis}
For a comprehensive understanding of our model, we further evaluate each module via ablation experiments on CIFAR-10+CIFAR-100 with 20\% and 40\% open-set noise. 

Table \ref{module_analysis} presents the following six experiments: (a) Without reweighting: we only change the weights of softmax loss on noisy samples by either assigning the same weights as clean samples (case $a_1$: $\gamma = 1$), or ignoring noisy samples (case $a_2$: $\gamma = 0$). (b) Without discriminative feature learning: we only remove the contrastive loss (the remaining model still has iterative noisy label detection and reweighted softmax loss modules). As for the detected noisy samples, we either remove them (case $b_1$: removing) or label them to a new class ``unknown" (case $b_2$: new class). (c) Without iterative noisy label detection: we either conduct detection only once at the first iteration of detection (case $c_1$: only once) or remove the detection module along with the reweighting on softmax loss (case $c_2$: no).

\begin{table}[!t]
\centering
\small
\caption{Accuracies (\%) on CIFAR-10+CIFAR-100 ($20\%$ \& $40\%$ open-set noise) after removing (w/o) each module from our model.}
\label{module_analysis}
\vspace{0.05in}
\begin{tabular}{l|cc}
\hline
\multirow{2}{*}{Method} & \multicolumn{2}{c}{CIFAR-10+CIFAR-100} \\ 
 & $20\%$ noise & $40\%$ noise \\ \hline
Our model & 81.96 & 79.28  \\ \hline
(a) w/o reweighting \\
{ }{ }-{}- case $a_1$: $\gamma = 1$ & 76.97 & 74.45  \\
{ }{ }-{}- case $a_2$: $\gamma = 0$ & 79.27 & 76.03 \\ \hline
(b) w/o discriminative learning  \\
{ }{ }-{}- case $b_1$: removing & 76.22 & 68.40\\
{ }{ }-{}- case $b_2$: new class & 78.34 & 73.11\\ \hline
(c) w/o iterative detection  \\
{ }{ }-{}- case $c_1$: only once & 77.52 & 70.31 \\
{ }{ }-{}- case $c_2$: no  & 76.17 & 63.50 \\ \hline
\end{tabular}
\vspace{-0.1 in}
\end{table}

Performance drops are observed in Table \ref{module_analysis} when any of the three modules is removed or replaced. The accuracy drop compared to our original model can be interpreted as the contribution of the module. Particularly, if discriminative feature learning (case $b_1$) or iterative noisy label detection (case $c_2$) is removed, the accuracy significantly decreases, which indicates that the two modules work jointly in an efficient way and can enhance each other. When the detected noisy samples are removed (case $b_1$) or re-labeled to a new class ``unknown" (case $b_2$), the accuracy also drops considerably, which proves that discriminative features forced by the contrastive loss are critical for accurate noisy label detection which further improves discriminative feature learning. From both cases in (a), we can see that reweighting is necessary for a proper handling of the detected noisy samples which may contain some clean samples around the decision boundary.

\subsubsection{Parameter and complexity analysis}\label{sec: parameters}
Moreover, we assess the influence of parameter $\eta$ in our model, which is used to balance the contrastive loss and the softmax loss. We test a series of $\eta \in [0.5, 1.5]$ on CIFAR-10+CIFAR-100 with 40\% open-set noise. Table \ref{eta_contrastive} shows that our model is not sensitive to the parameter $\eta$ as long as it lies in a comparable range to the weight of softmax loss on clean samples (its weight is 1). 

\begin{figure}[!t]
\centering
\small
\vspace{-0.06in}
\includegraphics[width=0.8\linewidth]{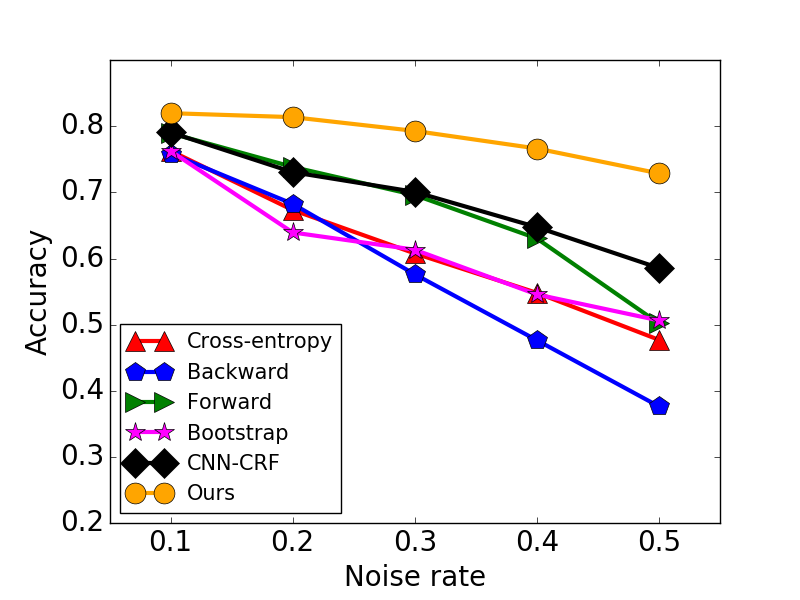}
\caption{Accuracies of different models on CIFAR-10+CIFAR-100 with different open-set noise rates.}
\vspace{-0.1in}
\label{noise_rate}
\end{figure}

We also compare our model with other baselines against different open-set noise rates on CIFAR-10+CIFAR-100. It can be seen in Figure \ref{noise_rate} that our model still performs the best under all noise rates, even at a high noise rate up to 50\%. The performance of baselines, however, decrease significantly as noise rate increases.

As for the complexity of our framework, the extra costs mainly lie on pcLOF computing. However, it is computed 1) within each class not on the entire data, 2) in parallel for different classes and 3) only every 10 epochs not each epoch. Moreover, pcLOF can be computed on GPU with 100X speed-up \cite{alshawabkeh2010accelerating}. The proportion of the time cost of pcLOF with respect to the training time is: $\frac{t_{pcLOF}}{t_{training}} \approx 2\%$. Thus, the computation of pcLOF is much less expensive. 

\begin{table}[!t]
\centering
\small
\caption{Accuracies (\%) of our model on CIFAR-10+CIFAR-100 ($40\%$ open-set noise) with different $\eta$. } 
\label{eta_contrastive}
\vspace{0.05in}
\begin{tabular}{c|cccccc}
\hline
$\eta$ & 0.5 & 0.7 & 0.9 & 1.0 & 1.3 & 1.5 \\ \hline
Ours & 76.08 & 79.04 & 79.36 & 79.77 & 79.08 & 77.80 \\
\hline 
\end{tabular}
\vspace{-0.1in}
\end{table}

\subsection{Experiments on ImageNet}\label{sec:imagenet}
From above experiments, we have demonstrated that our model achieves superior performance on a small dataset CIFAR-10 against both open-set and closed-set label noise. Here, we further present its capacity to handle large-scale datasets containing open-set noisy labels, \textit{i.e.}, ImageNet. Meanwhile, we also show that our model works effectively with different modern deep neural network architectures: ResNet-50 \cite{he2016deep} and Inception-v3 \cite{szegedy2016rethinking}.

\noindent\textbf{Experimental setup:}
Based on the ImageNet 2012 dataset \cite{deng2009imagenet} (1.3M images of 1000 classes), we generate an open-set noisy dataset by randomly taking 200 classes of images as clean data, which are then mixed with uniformly and randomly selected images from other 800 classes. Finally, we obtain a noisy dataset of $\sim$290k images with 200 classes and 20\% noise rate in each class. ResNet-50 and Inception-v3 networks are implemented in Keras \cite{chollet2015}. We train the networks with batch size 128 and initial learning rate 0.01, which is reduced by $1/10$ at the 30-th, 60-th and 80-th epoch. The training ends at the 100-th epoch. Several commonly used data augmentations are applied, \textit{i.e.}, $224 \times 224$ pixel random crops, horizontal random flips and scale data augmentation as in \cite{gross2016training}. All images are normalized by the per-color mean and standard deviation. We test the models using the ImageNet validation set of the 200 clean classes along with a single center crop ($224 \times 224$).

\noindent\textbf{Results:}
The baselines compared here are the same as that in Section \ref{sec:small_datasets}, and we report the Top-1 and Top-5 classification accuracy. The results can be found in Table \ref{tab:imagenet}. Again, our model outperforms other baselines with significant margins for ResNet-50 as well as Inception-v3 architectures. Meanwhile, we notice that some baseline models such as CNN-CRF and Forward also demonstrate certain improvements compared to Cross-entropy. This may because that some ImageNet classes are fine-grained and visually similar such that some open-set noisy images are ``closed-set" to some extent. In fact, this is in line with complex real-world situations, where a clear boundary between closed-set noisy labels and open-set noisy labels is often hard to draw. The superiority of our model on such datasets implies its advantages against real-world noisy data, where closed-set and open-set label noise may coexist.

\subsection{Experiments on real-world noisy dataset}
Finally, we assess our model on real-world noisy dataset, where noise type (closed-set or open-set) and noise rate are unknown. This is to demonstrate that our model can effectively make use of noisy data (web-search data) to learn accurate representation for real-world scenarios, \eg, no clean labeled data are available for a new domain.

\noindent\textbf{Experimental setup:}
We use Google image search engine to obtain a web-search image set of $\sim$1.2M images ($\sim$0.8k per class) using the query list from ImageNet \cite{deng2009imagenet}, SUN \cite{xiao2010sun} and NEIL \cite{chen2013neil} datasets, similar as in \cite{chen_iccv15}, but we do not remove images from the searched results. For conceptually clear queries, Google returns relatively clean results, however, for ``ambiguous" queries, the results are very similar to open-set label noise, \eg, `jasmine' query returns a plant or a cartoon character. The experimental setup and training are same as Section \ref{sec:imagenet}. For testing, however, this is a challenging task as there are no clean test data available. There are several ways to evaluate the learned features. Here we adopt a well defined classification task to achieve that. Specifically, we treat the learned network as a feature extractor by fixing all its layers except the softmax layer. We then train a softmax layer for classification using the CIFAR-100 training set. Because the optimization of the classification layer (\ie, multi-class logistic regression) is convex, the classification accuracy can be used to evaluate the learned features.

\noindent\textbf{Results:}
The classification accuracy on CIFAR-100 test set is reported in Table \ref{tab:web}. As shown, our model achieves the best performance compared to baselines for both ResNet-50 and Inception-v3. This indicates that our model can robustly train CNNs from real-world noisy dataset. Note that the noisy data are from web, thus our model can also be extended to webly supervised learning \cite{chen_iccv15}. 

In summary, empirical evidence has demonstrated that our model is a promising framework for learning from datasets with noisy labels (open-set and closed-set). Moreover, it can also be used for webly supervised learning. 

\begin{table}[!t]
\centering
\small
\caption{Accuracies (\%) of different models on the 200-class ImageNet with 20\% open-set noise. The best results are in \textbf{bold}.}
\vspace{1.5mm}
\label{tab:imagenet}
\begin{tabular}{l|cc|cc}
\hline
\multirow{2}{*}{Method}  
 & \multicolumn{2}{c|}{ResNet-50}  & \multicolumn{2}{c}{Inception-v3} \\
 & Top-1  & Top-5  & Top-1  & Top-5  \\	 \hline
Cross-entropy & 58.51 & 75.62 & 60.73 & 76.75  \\
Backward & 59.32 & 75.61 & 61.27 & 76.74 \\
Forward & 64.17 & 79.43 & 65.48 & 80.68 \\
Bootstrapping & 59.05 & 75.00 & 61.50 & 76.13 \\
CNN-CRF & 66.54 & 82.37 & 67.23 & 84.12  \\
Ours & \textbf{70.29} & \textbf{86.04} & \textbf{71.43} & \textbf{87.87} \\ 
\hline
\end{tabular}
\vspace{-0.02in}
\end{table}
\begin{table}[!t]
\centering
\small
\caption{Accuracies (\%) of different models trained on real-world noisy data (web-search data) and tested on CIFAR-100 test set. The best results are in \textbf{bold}.}
\vspace{1.7mm}
\label{tab:web}
\begin{tabular}{l|cc}
\hline
Method & ResNet-50 & Inception-v3 \\ \hline
Cross-entropy & 57.32 & 53.82 \\
Backward & 58.75 & 54.02 \\
Forward & 61.65 & 58.28 \\
Bootstrapping & 57.62 & 54.49\\
CNN-CRF & 63.94 & 60.47 \\
Ours & \textbf{67.90} & \textbf{64.21} \\
\hline
\end{tabular}
\vspace{-0.12in}
\end{table}

\section{Conclusions}
In this paper, we identified and investigated the open-set noisy label problem -- a more complex noisy label scenario that commonly occurs in real-world datasets. We proposed an iterative learning framework to address the problem with three powerful modules: iterative noisy label detection, discriminative feature learning, and reweighting. These modules are designed to benefit from each other and to be jointly improved over iterations. We empirically show that our model not only outperforms the state-of-the-arts for open-set label noise, but also effective for closed-set label noise, on datasets of various scales. 

{\small
\noindent \textbf{\em Acknowledgment.}
This work is supported by National Natural Science Foundation of China (No. 61771273), NSFC U1609220, NSF IIS-1639792 EAGER, ONR N00014-15-1-2340, Intel ISTC, Amazon AWS and China Scholarship Council (CSC). 
}
{
\small
\bibliographystyle{ieee}
\bibliography{egbib}
\nocite{liu2017deep}
\nocite{wang2017novel}
\nocite{sun2014deep}
\nocite{wang2017link}
\nocite{wang2016bernoulli}
}

\end{document}